# Metacarpal Bones Localization in X-ray Imagery Using Particle Filter Segmentation

Z. Bardosi, D. Granata, G. Lugos, A. P. Tafti, S. Saxena

IEEE Members

## Abstract

Statistical methods such as sequential Monte Carlo Methods were proposed for detection, segmentation and tracking of objects in digital images. A similar approach, called Shape Particle Filters was introduced for the segmentation of vertebra, lungs and hearts [1]. In this contribution, a global shape and a local appearance model are derived from specific object annotated X-ray images of the metacarpal bones. In the test data a unique labeling of the bone boundary and the background points and a manual annotation is given. Using a set of local features (Haar-like) in the neighborhood of each pixel a probabilistic pixel classifier is built using the random forest algorithm. To fit the shape model to a new image, a label probability map is extracted and then the optimal shape is obtained by maximizing the probability of each landmark with the Differential Evolution algorithm.

## 1. Introduction

The different modalities of medical imaging are the most important source of diagnostics in modern medicine. At the diagnostic step however, the massive amount of data generated by the imaging device has to be filtered to contain only the relevant information. The first step of this process i s usually the segmentation of different anatomical structures. The manual segmentation can be very time demanding, so there's an extensive research on finding automatic segmentation techniques. In case of complex anatomical shapes (like bone boundaries) model building is challenging. Lately, the use of machine learning algorithms in segmentation has gained focus, because of their sample based nature and generalization capabilities. Active Shape Models (ASM) are linear deformable statistical models of shapes defined by a given set of ordered landmarks. In the training step, a set of training shapes are presented to the algorithm. The training aligns these shapes to a common centroid, and minimizes the rotational variances. The resulting li near model contains a base (mean) shape, and a set of deformation basis vectors. In case of a new image, local image features are defined for each landmark, and the optimization iteratively deforms the shape to fit to an example of the object in a new image [2]. The classical ASM model uses gradient based optimization which usually fails to find the global optimum, and requires good starting parameters. Al so, the features used to define the "borderness" of the landmark points are very simple, and not sensitive enough in the case of X-ray images.

In this paper, we present an implementation of an extended version of the classical ASM by utilizing a better local "borderness" model learned statistically from the input dataset and have used the Differential Evolution [3] optimization method to find the optimum shapes. These two extensions increase the effectiveness of the algorithm on the dataset considerably.

The rest of the paper is arranged as follows. The proposed approach and related techniques come in Section 2. Section 3 shows the experimental setup as well as the experimental evaluations. Conclusions and possible direction for future work are presented in Section 4.

## 2. Method

In this part we begin with a short introduction to Active Shape Models (ASM) followed by differential evolution algorithm and random forest classifier. We then explain our proposed method for Metacarpal bones localization in X-ray imagery.

### 2.1. Active Shape Models

Since almost every digital images and in particular medical images include different components, the effectively measure and detect the existence of a specific object could be a difficult task. Using Active Shape Models it could be possible to detect and analyze complex components in digital images in which a statistical model of the shape of the particular object in the image will employ to detect the object. Further information and recent advances on ASM can be found in [4], [5].

### 2.2. Differential Evolution

The DE (Differential Evolution) algorithm is an iterative, heuristic population based stochastic function minimizer to solve real -parameterized problems. It is an improvement of the traditional evolutionary EA. In each iteration the algorithm uses a set of possible solution elements (population) to represent a subset of the search space. The series of the populations generated by each iteration of the algorithm are called generations.

The DE algorithm uses nature inspired search operators, such as 1) Mutation: to introduce noise on the population members, 2) Crossover: to generate members by mixing the properties of existing members, and 3) Selection: to filter out the unfit members of the population [3].

In order to find the optimum, at each iteration:
1) New population members are generated (using mutation and crossover).
2) The best members of the population are selected (using a fitness-function).

There are several types of strategies that can be used as mutation, cross-over and selection operators. Each configurations of these operators define a strategy for the optimizer. On different problem types different strategies can be effective. Though the algorithm i s quick and simple, it i s heavily dependent on the used strategy and on the refinement of parameters (like the mutation rate F, and crossover probability CR).

Fundamental basics on differential evolution as well as its application in computer vision problems can be found in [6], [7], [8], [9].

## 2.3. Random Forest Classification

Random forest is a type of machine learning classification strategies which triggers by constructing a group of decision trees at training time. In the other word, a random forest grows several classification trees. To classify a new item from an input vector, we put the input vector downward each of the trees in the forest. Therefore, each tree can give a classification and it votes for the class. Then, the forest will pick up the classification which has the most votes among all the trees in the forest. Readers interested in random forest classification are referred to [10], [11] for detailed information.

## 2.4. The Proposed Algorithm

In the first step of the algorithm, a linear shape distortion model is learned from the training dataset, similarly to the ASM method. Figure 1 shows the application of an ASM model on knee X-ray image. By finding the principal directions of variance in the shape space, the dimensionality of the model can be greatly reduced. The standard approach is to maintain approximately 95% of the variance present in the training dataset. The result is a generative shape model that can create different shapes from a low-dimensional linear model. Our original shape dimensionality is 128, since we have 2 coordinates, and 64 landmarks. With this method the dimensionality of the shapes reduced to 8 (in our test) shape deformation dimensions and 5 transformation dimensions ( translation in x,y, scaling in x,y, and rotation around z: Tx, Ty, Sx, Sy, Rz ). This 13 dimensional shape model can generate a wide variety of shapes, that still resemble the training dataset, but can also generate new shapes, that were previously unseen.

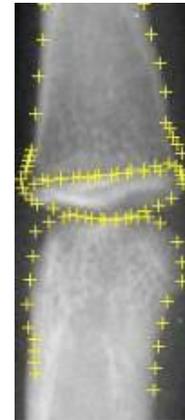

Figure 1: Active Shape Model on knee X-ray image.

In the second step, we build a pixel based classifier to distinguish between two classes of pixels. We use the manual classification samples given in the database to train this classifier. The database contains mask i mages, that assign a label to each pixel on each image. The border of the interesting metacarpal bone structure is assigned a 'positive' label, while the rest of the image pixels belong to the 'background' class. Readers interested in manual or automatic classification using machine learning approaches are refereed to [12], [13] for other techniques.

In order to differentiate between the two pixel classes, we use a set of local i mage features centered around the pixel-of-interest. Our image features were motivated by the work of Viola Jones [14] that uses a large set of local features for face detection. The authors in this paper have used a set of i mage features that resemble the Haar-wavelet basis functions in 2D. The main advantage of these features i s that they can be very quickly evaluated on the integral images. In our implementation we have used 14 preselected Haar-like feature that were evaluated i n the neighborhood of the central pixel. Windows sizes between 3×3 - 9×9 were used. The responses of the detectors were collected for every 'positive' boundary pixel, and a set of random background pixels were used to have the negative or 'background' class samples. In the next step, we have trained a random forest classifier [10] namely the TreeBagger implementation in MATLAB, to learn the difference between a boundary and a non-boundary pixel. Figure 4 shows the basic structure of a random forest classifier. With an ensemble of 32 decision trees we have learned a classifier that renders a probability value to each pixel on which it is evaluated. This probability corresponds to the likelihood that the given pixel belongs to the 'border' class.

By evaluating the random forest decision method on every pixel of every test image we have gathered a probability map for each test image. The last step of our algorithm uses a DE optimizer to optimize the shape and transformation parameters to fit a shape on the probability map. At each shape and transformation parameter, the 2D vector of the shape landmarks are generated. Then, at each landmark the probability map is sampled to check the likelihood that the landmark is positioned on the border of the bone. The likelihood of every landmark is then averaged. This renders a probability to the shape as a whole. The task of the optimizer is to find the parameterization for which the shape likelihood is maximal. In order to find the shape and transformation parameters that yield the maximum probability on the calculated probability map, we have used the Differential Evolution algorithm.

The general pipeline and the different steps of our proposed method are shown in Figure 2 and Figure 3.

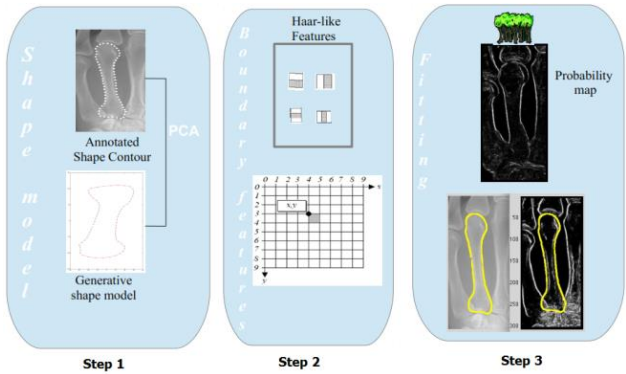

Figure 2: The pipeline of the proposed method.

We have used image dependent bounds on the transformation parameters to align the shapes at around the center of the images, with about 100 pixel translations in each direction. The allowed scale factor was about 0.8-1.5 compared to the aligned shape size, and we have allowed a +- 45 degrees of rotation freedom around the centroid of the shape. To constrain the deformation parameters of the shape model we have used the $c_i$ parameter calculated from the eigenvalues of the most important principal components (for $i = 1..8$). We allowed a +-3* $c_i$ in each deformation dimension.

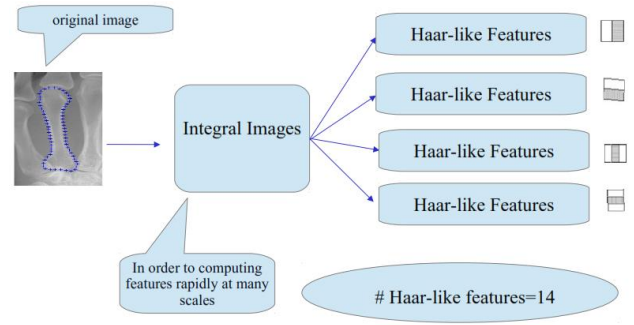

Figure 3: Using Haar-like features.

## 3. Experimental Results

In this section we further analyze our proposed approach by applying it on image dataset contained 50 grayscale X-Ray images of metacarpal bone radiographs. The resolution of the images was approximately 300×150. Each image was annotated using 64 landmark positions defined on the boundary of the middle metacarpal bone.

The shape learning process resulted in a 8 dimensional parameter space for deformations. The associated eigenvalues of the PCA method are shown in Figure 4. The vertical direction shows the variance associated with the given eigenvector, the horizontal axis shows the index of the eigenvectors in decreasing order of importance. We have selected the first 8 (most relevant) eigenvectors to be included into our shape model.

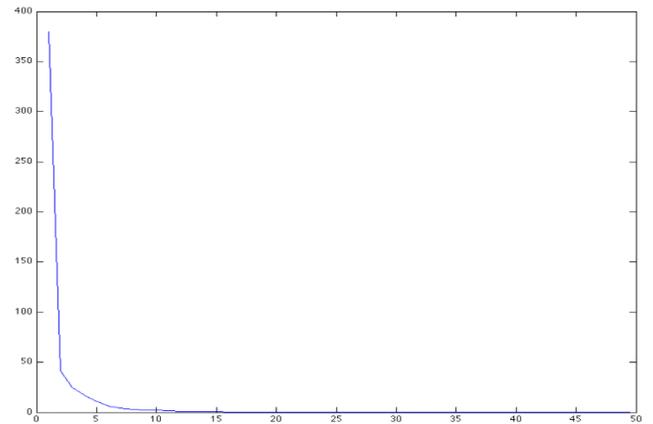

Figure 4: Eigenvalues associated with the shape deformation components.

Some sample shapes generated from the shape model, 1) by applying the first principal component to the mean shape with different weights (Figure 5), and 2) by applying different transformations (Tx, Ty, Sx, Sy, Rz) to the shape (Figure 6).

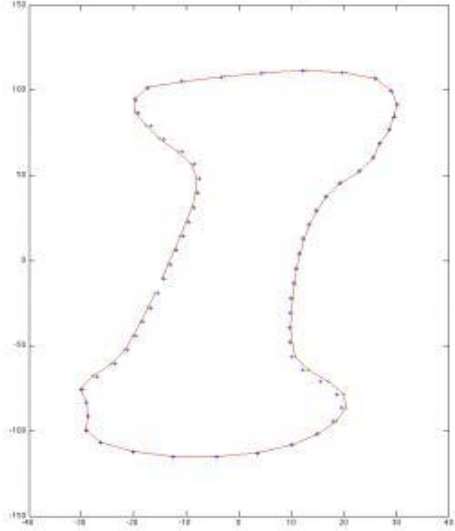

Figure 5: Applying the first principal component to the mean shape.

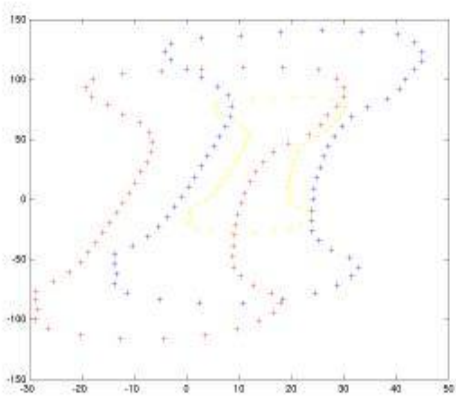

Figure 6: Applying different transformation to the shape.

In the local pixel learning step, we have built a pixel-based classifier. We have used 4 times as many negative samples to avoid false positive classifications in the difficult image regions. From the 30 training images we've gathered approximately 65000, 14 dimensional training samples. Each sample was assigned to a label coming from the bone boundary mask. A random forest classifier was trained using 32 trees to classify between the bone-border and the background pixels.

Figure 7 shows one of the resulting probability maps overlayed by two different shapes. The figure shows how different shapes are generated onto the probability map in order to evaluate their fitness to be the best boundary shape.

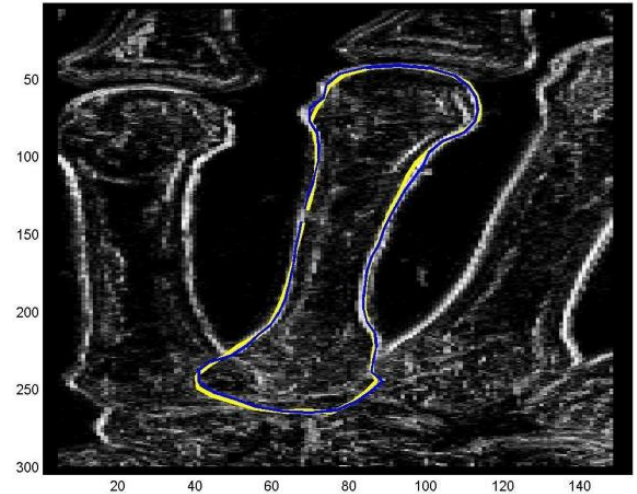

Figure 7: Result.

In the optimization step, we have used the standard MATLAB implementation of the DE solver. We've used a "minimum cost function" version 1.0 - probability (shape) for the optimizer in order to turn the task into a minimization problem. Some tweaking of the optimizer parameters was required. After a few trial-and-error we have found that the optimizer works best when we use the following strategies and parameterization:

("DE/rand/1/bin", F = 0.5, and CR = 0.75)

This strategy selects the members of the current population randomly for mutation. The crossover strategy uses the binomial-crossing. F is the mutation rate, CR is the crossover rate.

A small video demonstration shows the optimization running on one of the probability maps [15]. At the initial steps the population is far away from the optimum and changes rapidly. In the latter iterations the optimum is nearly found, and the model changes only in small steps. At the end of the video the manual annotation is al so shown with blue line. In the final test we have constrained to optimizer to stop at 95% of probability or after 1500 iteration. The first termination condition was never reached. It is possible that with further iterations the results could have been a bit better. We have used 20 i mages that were not seen in the training test as a test sample, and we have compared the position of each annotated landmark with the appropriate landmark from the result of the optimization method. The mean square errors on the landmark distances were calculated (Figure 8).

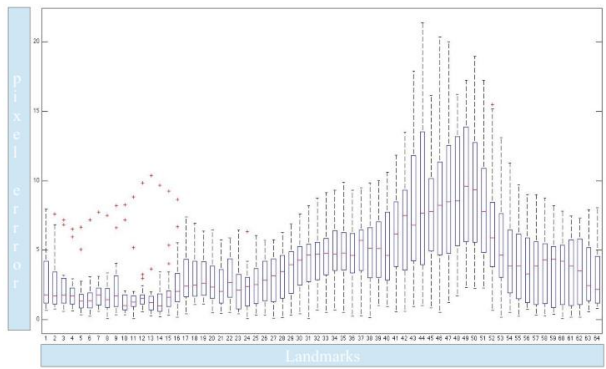

Figure 8: Distribution of the landmark errors.

Notably the error rate of the landmarks 41-51 is much higher than the error on the rest of landmarks. This is due to the fact that these landmarks are positioned in a region of the image, where the X-ray images contain multiple shadows and false boney edges.

## 4. Conclusion and future work

The proposed model presents an accurate representation of the spatial shape of the object of interest, but the accuracy of the results is limited due to shadow/overlap artifacts generated by the X-ray imaging. The pixel based local classifier has rendered higher probabilities to the false boundary than to the manually annotated boundary. In this region, the optimizer can only rely on the shape constrains and not the local probability information.

Another issue is that the shape model can only fit perfectly on the shapes that are in the span of the principal deformation components. If the shape training set contains too much variability it will over fit. If it does not contain enough variability, some important shape configurations will never be generated by the optimizer. The following improvements for Shape Particle Filters are expected to improve performance and accuracy:

- Investigation of the optimal selection of feature types for specific target objects.
- Using several landmark-based statistic tool analysis for improving the current parameters and i n order to minimize the cost function.
- Using a probability atlas to improve the quality of the segmentation in the problematic regions.
- Incorporating additional elasticity into the shape model during the Differential Evolution process to refine newly generated shapes and push them towards the shape to segment similar to ASMs. This would lead to a combination of the advantages of both methods.
- Low-level and image processing techniques will improve the accuracy of the work. There are several sources in X-ray imagery that produce different kind of noises which consequently decrease the reliability and accuracy of the work. Modern digital image processing techniques such as noise reduction [16], image registration [17] may apply to increase the fidelity of the proposed system.
- The value of any research project might be evaluated by how much it is effective in real applications. An eligible improvement would be expanding the application areas from biomedical imaging, to material, mechanical and other engineering fields [18] Designing and developing open-source and N-Tier reusable software application [19], [20], [21] for particle filter segmentation will call for bioimaging software reusability.
- A small amount of independent (i.e. Gaussian, 1-2 pixel wide) noise added to the individual landmarks would extend the search space of the optimizer with shape elements that are not in the span of the PC components.